\documentclass[conference]{IEEEtran}
\IEEEoverridecommandlockouts
\usepackage{cite}
\usepackage{amsmath,amssymb,amsfonts}
\usepackage{algorithmic}
\usepackage{graphicx}
\usepackage{textcomp}
\usepackage{xcolor}
\usepackage{amssymb}
\usepackage{latexsym}
\usepackage{multirow}
\usepackage{enumitem}
\usepackage{subfigure}  
\usepackage{booktabs}
\usepackage{bbding}
\usepackage{pifont}

\def\BibTeX{{\rm B\kern-.05em{\sc i\kern-.025em b}\kern-.08em
    T\kern-.1667em\lower.7ex\hbox{E}\kern-.125emX}}
\begin{document}

\title{Slightly Shift New Classes to Remember Old Classes for Video Class-Incremental Learning}
\author{\IEEEauthorblockN{Jian Jiao, Yu Dai, Hefei Mei, Heqian Qiu, Chuanyang Gong, Shiyuan Tang, Xinpeng Hao, Hongliang Li}
\IEEEauthorblockA{\textit{University of Electronic Science and Technology of China}\\
{jij2021, ydai, hfmei}@std.uestc.edu.cn, hqqiu@uestc.edu.cn, {cygong, sytang, xphao}@std.uestc.edu.cn, hlli@uestc.edu.cn}
}

\maketitle

\makeatletter
\def\ps@IEEEtitlepagestyle{%
\def\@oddfoot{\mycopyrightnotice}%
\def\@evenfoot{}%
}
\def\mycopyrightnotice{%
\gdef\mycopyrightnotice{}
}

\begin{abstract}
Recent video class-incremental learning usually excessively pursues the accuracy of the newly seen classes and relies on memory sets to mitigate catastrophic forgetting of the old classes. However, limited storage only allows storing a few representative videos. So we propose SNRO, which slightly shifts the features of new classes to remember old classes. Specifically, SNRO contains Examples Sparse(ES) and Early Break(EB). ES decimates at a lower sample rate to build memory sets and uses interpolation to align those sparse frames in the future. By this, SNRO stores more examples under the same memory consumption and forces the model to focus on low-semantic features which are harder to be forgotten. EB terminates the training at a small epoch, preventing the model from overstretching into the high-semantic space of the current task. Experiments on UCF101, HMDB51, and UESTC-MMEA-CL datasets show that SNRO performs better than other approaches while consuming the same memory consumption.
\end{abstract}


\section{Introduction}

Human activity recognition has been widely studied in recent years, which usually training models with all classes in one stage\cite{carreira2017quo,karpathy2014large,lin2019tsm,tran2015learning,feichtenhofer2019slowfast,li2022mvitv2} in a large-scale video dataset. 
However, in many real-world application scenarios, due to privacy protection or technical limitations, model training is often divided into multiple tasks where different classes appear in sequence.
Since the model tends to heavily over-fit the classes available in the current task, causing performance deterioration on those previously seen, this problem is called catastrophic forgetting\cite{mccloskey1989catastrophic}.
The method used to solve this question is class-incremental learning, which aims to maintain the model's performance in a sequence of independent tasks.

\begin{figure}
	\centering  
      \includegraphics[width=0.99\linewidth]{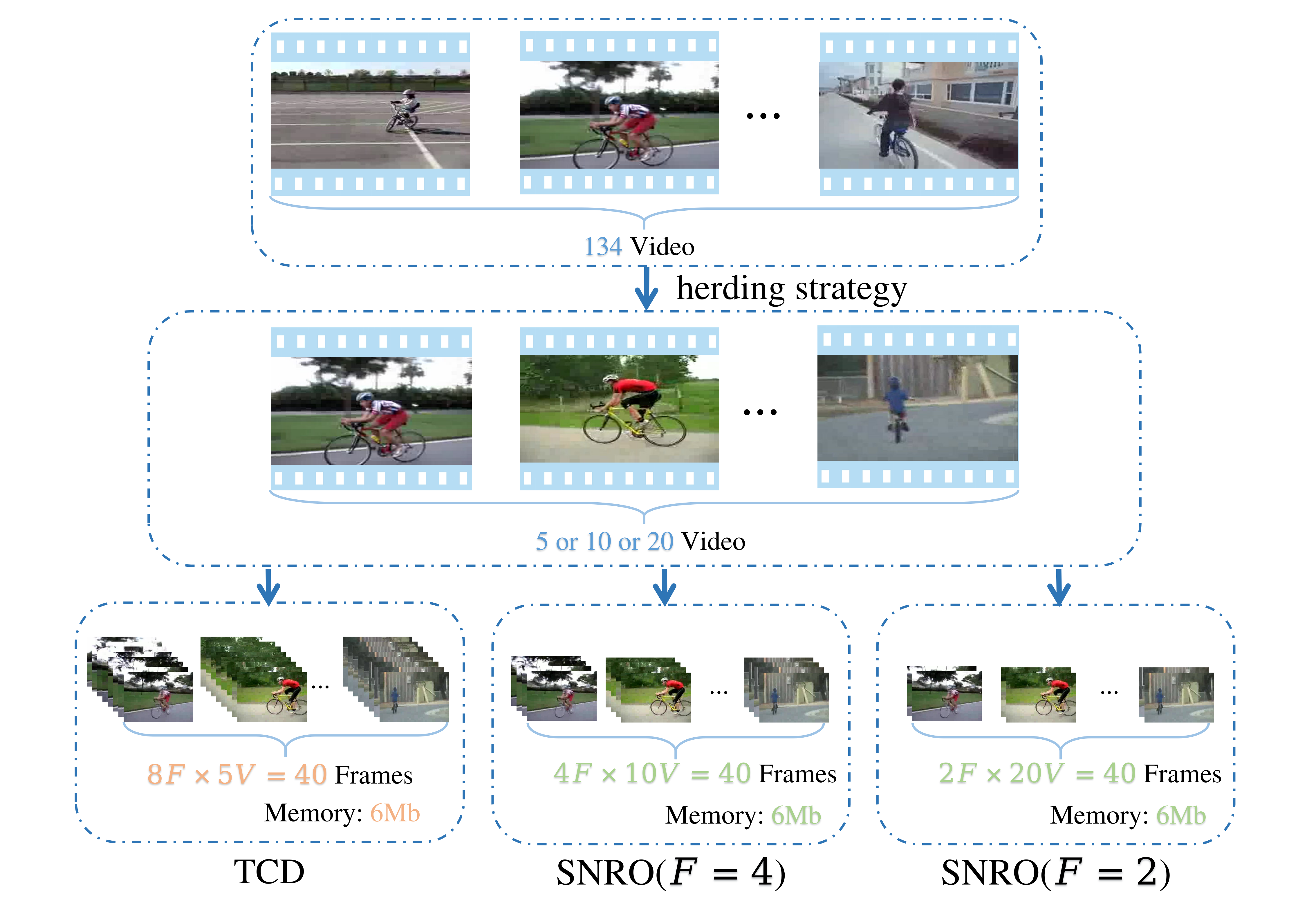}
	\caption{Analysis for memory consumption. $xF\times yV=zMb$ means that sampling $x$
frames for each video from $y$ different videos to store for each class. Assume the resolution of frame is $3\times 224\times 224$, and the total memory consumption is $zMbytes$}
	\label{intruduction}
	\vspace{-1.0em}
\end{figure}

\begin{figure*}
    \centering  
    \centerline{\includegraphics[width=0.95\linewidth]{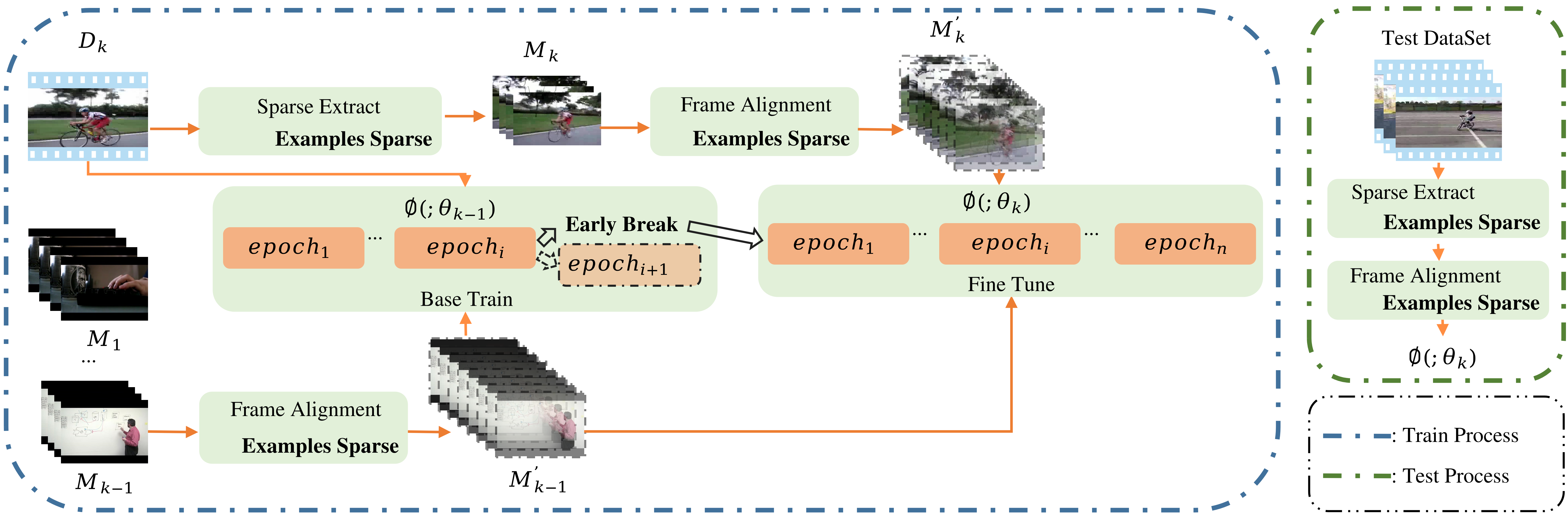}}
    \caption{llustration of the proposed SNRO framework.  Note that we also used Examples Sparse in the testing phase}
    \label{totalArch}
    \vspace{-1.0em}
\end{figure*}

Many class-incremental learning methods have achieved remarkable performance in the image domain\cite{castro2018end,douillard2020podnet,hou2019learning,li2017learning,wu2019large}.
They demonstrate that the memory sets of old classes can relieve catastrophic forgetting, with more examples in previous tasks stored,  less catastrophic forgetting happens. 
Existing video class-incremental learning methods\cite{villa2022vclimb,park2021class} also verify this.
At the end of every task, they manage to select a small count of representative videos and then construct a memory set from these videos shown as Fig. \ref{intruduction}.
For these representative videos: 
vCLIMB\cite{villa2022vclimb} extracts 8-16 frames for each video and down-sample these frames to store, leading to non-negligible memory overhead. 
TCD\cite{park2021class} extracts 8 frames for each video and reaches a remarkable performance based on knowledge distillation, but still not relieve the high memory overhead. 

Besides, the existing video class-incremental approaches excessively pursue the recognition accuracy for the current new classes in each incremental task. They believe that the model should first fully fit the high-semantic features of the new classes, and then consider how to maintain the features of the previous classes.
However, lots of features of old classes have been overwritten by new classes in this process, and have no efficient ways to retrieve these lost features.

To remedy the above weaknesses, we present the proposed SNRO.
SNRO significantly alleviates the catastrophic forgetting of old classes at the cost of slightly drop the performance of the current new classes, thereby improving the overall recognition accuracy.
SNRO consists of two essential parts: Examples Sparse and Early Break.
Examples Sparse firstly performs Sparse Extract on representative videos of old classes and saves those sparse frames, storing a larger memory set under the same memory consumption. 
Frame Alignment is then used to align sparse frames with the network input. Since the sparse frames contain less spatio-temporal information than the original frames, the network can be prevented from excessively extending to the high semantic space.
Early Break is used to terminate training earlier during each incremental task, trying to give up the high performance for current new classes of the model which is usually accompanied by over-fit.

The main contributions of our work could be summarized as follows:

\begin{itemize}[itemsep= 0 pt,topsep = 0 pt]
\item We use Sparse Extract to save larger memory sets with the same space consumption as other methods, effectively alleviating the forgetting of old classes.

\item We use Frame Alignment to reduce the spatio-temporal information of video representation and use Early Break to prevent the model from over-stretching to newly seen classes. By slightly dropping the performance of the current task, we greatly improve the performance of previous tasks.

\item The experiments on UCF101 dataset \cite{soomro2012ucf101}, HMDB51 dataset\cite{kuehne2011hmdb} and UESTC-MMEA-CL dateset\cite{xu2023towards} demonstrate the effectiveness of the SNRO. 

\end{itemize}

\section{Proposed Method}

This section presents the framework of SNRO as shown in Fig.~\ref{totalArch}, which mainly consists of two crucial components: Examples Sparse and Early Break. 
Examples Sparse decimates videos of old classes at a lower sample rate and uses interpolation to align frames in the future training stage.
Early Break terminates the training at a small epoch when in the incremental training stage.

\subsection{Problem formulation}
Video class-incremental learning aims at training a model $\Phi(;\Theta)$ with parameter $\Theta$ while the tasks  $\{T_1,T_2,\cdot\cdot\cdot,T_k,\cdot\cdot\cdot\}$ arrive in a sequence.
Each task $T_k$ has its exclusive dataset $D_k$ which contains several classes that have never been seen in previous tasks, which means $(D_1\cup\cdot\cdot\cdot\cup D_{k-1})\cap D_k=\varnothing$.
The training of task $T_k$ could be divided into two stages: Base Train and Fine Tune.
According to existing works, we make a memory set $M_k$ for dataset $D_k$ when finished task $T_k$'s Base Train, and provide all memory sets to  $T_k$'s Fine Tune and  subsequent tasks.

\subsection{Examples Sparse}
We refer to the standard protocol of video class-incremental method\cite{park2021class} which is based on memory-replay strategy and knowledge distillation. 
We realize that rich context information will cause the model to stretch towards high-semantic space, leading to heavily relying on high-semantic features to interface. 
This kind of stretch is helpful in a single classification task but is disastrous in the class-incremental task. 
In the process of continuous training, high-semantic features are easily overwritten by new classes with sufficient samples, while the model has not learned enough ability to classify based on low-semantic features, which leads to the seriously forgotten of  old classes.

The meaning of Examples Sparse lies in this. Specifically, Examples Sparse contains two parts: Sparse Extract and Frame Alignment.
In task $T_k$'s Base Train stage, $(M_1\cup\cdot\cdot\cdot\cup M_{k-1})\cup D_k$ are accessible to update the model $\Phi(;\Theta_{k-1})$ and use Sparse Extract to establish $D_k$'s memory set $M_k$. 
We first select a subset of video $C_k$ by herding strategy\cite{rebuffi2017icarl} from $D_k$ after task $T_k$'s Base Train. 
Assume that the action recognition network needs $F$ frames of a video to perform classify, thus given a video $C_{k}^i\in C_k$, TCD\cite{park2021class} extract $F$ frames uniformly, while we only sample $\Bar{F}(\Bar{F}=\frac{F}{2} or \frac{F}{4})$ frames uniformly to build memory set.
In ${T_k}$'s Fine Tune stage, memory sets $(M_1\cup\cdot\cdot\cdot\cup M_{k})$ are accessible to update the model $\Phi(;\Theta_{k})$ trained in Base Train stage.
Frame Alignment is used to align the memory set with $\Bar{F}$ frames to $F$ frames the network requires. 
For example, if $\Bar{F}=\frac{F}{2}$, given extracted frames ${I_k^{i1},I_k^{i2},\cdot\cdot\cdot,I_k^{i{\frac{F}{2}}}}$, to avoid additional computational consumption, we choose Uniform Interpolation:
\begin{equation}
{I_k^{i1},\frac{I_k^{i1}+I_k^{i2}}{2},I_k^{i2},\cdot\cdot\cdot,I_k^{i{\frac{F}{2}}}}
\label{Uniform Interpolation}
\end{equation}
or Repeated Interpolation:
\begin{equation}
{I_k^{i1},I_k^{i1},I_k^{i2},\cdot\cdot\cdot,I_k^{i{\frac{F}{2}}}}
\label{Repeated Interpolation}
\end{equation}
to align them to $T$ frames.
in fact, Frame Alignment is also required in $T_k$'s Base Train stage for it uses memory sets too. 
Notice that we do not perform Examples Sparse on $D_K$ in $T_K$'s Base Train which may heavily sacrifice the performance of the new classes. $D_K$ firstly helps the model achieve decent performance on new classes, and a part of its high-semantic features are then shifted to low-semantic features in Fine Tune stage due to its sparse $M_k$. 

\begin{figure}
	\centering  
	\subfigure[GradCAM maps in TCD]{
		\includegraphics[width=1\linewidth]{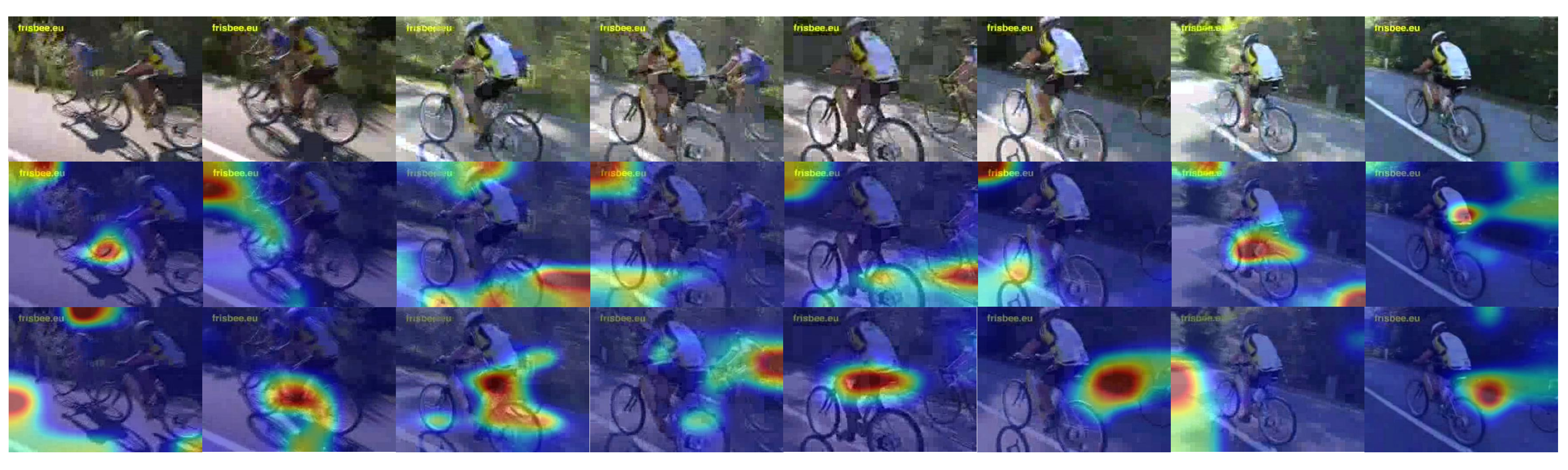}}
           \label{TCD-6-8}
      \vspace{-0.5em}    
	\subfigure[GradCAM maps in SNRO]{
		\includegraphics[width=1\linewidth]{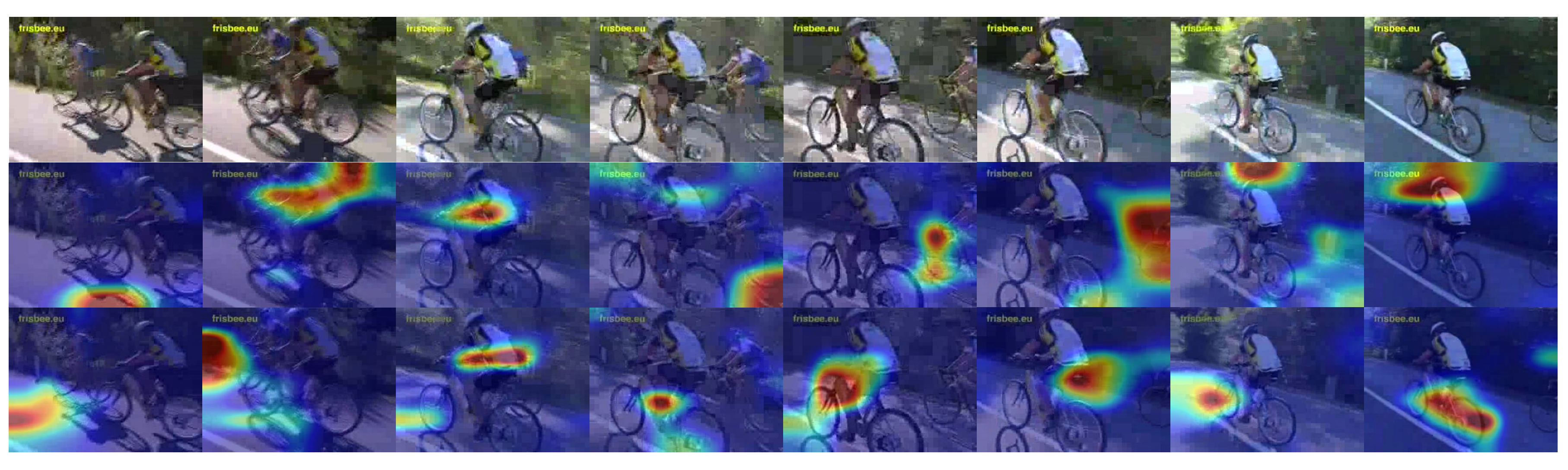}}
           \label{SNRO-6-8}
    \vspace{-1.0em}
	\caption{$(a)$ and $(b)$ show visualization of GradCAM maps in different tasks of a video labeled "Biking" in TCD and SNRO. This class appears in task $T_6$. The first row of $(a)$ and $(b)$ are raw frames. The second and the third row are their corresponding GradCAM maps at the end of task $T_6$ and task $T_8$. SNRO converges worse than TCD on the bicycle's feature at the end of $T_6$, which means "Shift New Classes". But is better than TCD at the end of $T_8$, which means "Remember Old Classes."}
	\label{CAM}
      \vspace{-1.0em}
\end{figure}

Examples Sparse ensures we build larger memory sets consuming the same space as TCD.
And using $\frac{F}{2}$ frames to represent a video contains less spatio-temporal information than using $F$ frames, it effectively prevents the network from over-stretching to high-semantic spaces, which allows preserving more low semantic features in future incremental tasks as shown in Fig.~\ref{CAM}.$(a)$ and Fig.~\ref{CAM}.$(b)$.
Since Examples Sparse makes the network more dependent on low-semantic features for classification, we also do Examples Sparse in the interface stage.

\subsection{Early Break}
To better prevent the model from over-fit to new classes with sufficient samples in task $T_k$, which is often caused by deploying too many training epochs\cite{shin2017continual,yoon2017lifelong,kirkpatrick2017overcoming}, we use Early Break to terminate training.
Under multiple experiments, we find in the incremental tasks $T_k (k>0)$, the model sticks in over-fit just in a few epochs, usually 5-10. Continuing to iterate in this sense causes the model to severely over-fit the new class, losing many features learned in previous tasks. 
Specifically, the initial task $T_0$ is fixedly trained for $N$ epochs, and its highest training average accuracy is set as the threshold in the incremental task $T_k(k>0)$'s Base Train stage. $T_k$'s Base Train can train up to $\frac{N}{2}$ epochs, and if the threshold is reached during this period, the training will be exited immediately. 
Note that we don't apply Early Break to Fine Tune stage, because it uses the memory sets of all seen classes with the same sample numbers.

\section{Experiments}

\begin{table*}
    \centering
    \begin{tabular}{ccccccccccc}
        \hline
        Dataset & \multicolumn{6}{c}{UCF101} & \multicolumn{4}{c}{HMDB51} \\
        \cmidrule(lr){1-1} \cmidrule(lr){2-7} \cmidrule(lr){8-11}
        Num. of Classes & \multicolumn{2}{c}{$10\times5$ stages} & \multicolumn{2}{c}{$5\times10$ stages} & \multicolumn{2}{c}{$2\times25$ stages} & \multicolumn{2}{c}{$5\times5$ stages} & \multicolumn{2}{c}{$1\times25$ stages} \\
        Classifier & CNN & NME & CNN & NME & CNN & NME & CNN & NME & CNN & NME \\
        \cmidrule(lr){1-1} \cmidrule(lr){2-3} \cmidrule(lr){4-5} \cmidrule(lr){6-7} \cmidrule(lr){8-9} \cmidrule(lr){10-11}
        Finetuning & 24.97 & - & 13.45 & - & 5.78 & - & 16.82 & - & 4.83 & - \\
        LwFMC\cite{li2017learning} & 42.14 & - & 25.59 & - & 11.68 & - & 26.82 & - & 16.49 & - \\
        LwM\cite{dhar2019learning} & 43.39 & - & 26.07 & - & 12.08 & - & 26.97 & - & 16.50 & - \\
        iCaRL\cite{rebuffi2017icarl} & - & 65.34 & - & 64.51 & - & 58.73 & - & 40.09 & - & 33.77 \\
        UCIR\cite{hou2019learning} & 74.31 & 74.09 & 70.42 & 70.50 & 63.22 & 64.00 & 44.90 & 46.53 & 37.04 & 37.15 \\
        PODNet\cite{douillard2020podnet} & 73.26 & 74.37 & 71.58 & 73.75 & 70.28 & 71.87 & 44.32 & 48.78 & 38.76 & 46.62 \\
        TCD\cite{park2021class} & 74.89 & 77.16 & 73.43 & 75.35 & 72.19 & 74.01 & 45.34 & 50.36 & 40.47 & 46.66 \\
        \cmidrule(lr){1-1} \cmidrule(lr){2-3} \cmidrule(lr){4-5} \cmidrule(lr){6-7} \cmidrule(lr){8-9} \cmidrule(lr){10-11}
        SNRO & \textbf{78.96} & \textbf{77.76} & \textbf{77.60} & \textbf{76.95} & \textbf{76.84} & \textbf{76.21} & \textbf{48.65} & \textbf{52.10} & \textbf{46.40} & \textbf{49.38} \\
        \hline
    \end{tabular}
    \vspace{0.5em}
    \caption{Comparison with the state-of-the-art approaches on UCF101 and HMDB51. }
    \label{tab:ucf and hmdb}
    \vspace{-2.0em}
\end{table*}

\begin{table}
    \centering
    \begin{tabular}{ccccccc}
        \hline
        Num. of Classes  & \multicolumn{2}{c}{$8\times4$ stages} & \multicolumn{2}{c}{$4\times8$ stages} & \multicolumn{2}{c}{$2\times16$ stages}\\
        Classifier & CNN & NME & CNN & NME & CNN &  NME \\
        \cmidrule(lr){1-1} \cmidrule(lr){2-3} \cmidrule(lr){4-5} \cmidrule(lr){6-7}
        Finetuning & 53.19 & - & 29.3 & - & 17.3 & - \\
        LwM\cite{dhar2019learning} & 51.6 & - & 40.4 & - & 21.9 & - \\
        EWC\cite{kirkpatrick2017overcoming} & 72.04 & - & 51.6 & - & 31.8 & - \\
        iCaRL\cite{rebuffi2017icarl} & 75.84 & - & 70.44 & - & 69.15 & -\\
        TCD\cite{park2021class} & 83.47 & 85.62 & 78.23 & 79.05 & 76.85 & 77.13 \\
        \cmidrule(lr){1-1} \cmidrule(lr){2-3} \cmidrule(lr){4-5} \cmidrule(lr){6-7}
        SNRO & \textbf{84.68} & \textbf{85.81} & \textbf{82.38} & \textbf{82.48} & \textbf{80.52} & \textbf{79.60} \\ 
        \hline
    \end{tabular}
    \vspace{0.5em}
    \caption{Comparison with the state-of-the-art approaches on UESTC-MMEA-CL. }
    \label{tab:uestc}
    \vspace{-1.0em}
\end{table}

\noindent
\textbf{Datasets.} \enspace We evaluated SNRO on three action recognition datasets, UCF101\cite{soomro2012ucf101}, HMDB51\cite{kuehne2011hmdb} and UESTC-MMEA-CL\cite{xu2023towards}. 
UCF101 dataset contains 13.3K videos from 101 classes.
We train the model on 51 classes in the initial task, and the remaining 50 classes are divided into groups of 10, 5, and 2 classes for each incremental task.
HMDB51 dataset contains 6.8K videos from 51 classes.
We train the model on 26 classes in the initial task, and the remaining 25 classes are divided into groups of 5 and 1 classes for each incremental task.
UESTC-MMEA-CL dataset contains 6.4K videos from 32 classes.
Follow the setting of\cite{xu2023towards}, we divide 32 classes into 4,8, and 16 groups, each group has 8,4, and 2 classes, and send groups to the corresponding task sequentially. 

\noindent
\textbf{Implementation Details.} \enspace We employ TSM\cite{lin2019tsm} as our backbone and follow its data pre-processing procedure. 
The proposed SNRO is based on TCD\cite{park2021class}.
For UCF101 dataset and UESTC-MMEA-CL dataset, we train a ResNet-34 TSM with a batch size 32.
For HMDB51 dataset, we train a ResNet-50 TSM with a batch size of 12.These settings are consistent with TCD.
we set $\Bar{F}=\frac{F}{4}$ which $F=8$ in the Examples Sparse strategy for UCF101, and set $\Bar{F}=\frac{F}{2}$ for HMDB51 and UESTC-MMEA-CL because their longer video lengh. 

In task $T_0$'s Base Train of each experiment, we train $N=50$ epochs and record the best performance as the Early Break's threshold in the incremental task $T_k(k>0)$.
Each task's Fine Tune is fixed to $30$ epochs.
For initialization, we use the ImageNet-pretrained weights instead of Kinetics-pretrained weights, preventing pretrained weights from already containing the information of unseen classes\cite{villa2023pivot}.
We run our experiments using three random class orders(random seed is set to 1000,1993 and 2021) and report their final average performance of CNN and NME.

\subsection{Comparison with State-of-the-art Results}
We compare the proposed SNRO with existing class-incremental learning approaches\cite{park2021class}.
For a fair comparison, we use the same exemplar memory size of $6Mb$ for each class.

To demonstrate that SNRO greatly improves the performance of old classes, we report the final average accuracy and average forgetting\cite{chaudhry2018riemannian} of SNRO and other methods on different datasets.
Assume $a_{k,j} (j\leq k)$ denotes the CNN or NME accuracy evaluated on the test set of task $T_j$ after incremental task $T_k$, then the average accuracy on task $T_k$ is defined as:
\begin{equation}
{ACC_k=\frac{l}{k} \sum_{j=1}^k a_{k,j}}
\label{average accuracy}
\end{equation}
The forgetting  $f_{k,j}$ is defined as the CNN or NME forgetting on task $T_j$ after incremental task $T_k$ which formulated as:
\begin{equation}
{f_{k,j}=\max _{l \in j, \cdot\cdot\cdot, k-1} a_{l,j}-a_{k,j}, \quad \forall j < k}
\label{forgetting}
\end{equation}
The average forgetting on task $T_k$ is defined as:
\begin{equation}
{FOR_k=\frac{l}{k-1} \sum_{j=1}^{k-1} f_{k,j}}
\label{average forgetten}
\end{equation}

Table \ref{tab:ucf and hmdb} reports the final task's average accuracy on UCF101 and HMDB51, which shows that SNRO outperforms other methods under different configurations in terms of all CNN and NME.
We compared SNRO with recent methods on UESTC-MMEA-CL in Table \ref{tab:uestc}, SNRO sets new state-of-the-art performance, which means it is also more effective on the large-scale dataset.

Table \ref{tab:forgetten} shows that the final task's average forgetting of SNRO is lower than the state-of-the-art methods on every dataset, meaning SNRO achieves better performance by reducing the forgetting of old classes, rather than pursuing the high performance of new classes.
\vspace{-0.25em}

\subsection{Ablation Study}
In this section, we analyze the effectiveness of SNRO by presenting ablation studies on UCF101 dataset.

\begin{table}
    \centering
    \begin{tabular}{ccccccc}
        \hline
        Dataset & \multicolumn{2}{c}{UCF101} & \multicolumn{2}{c}{HMDB51} & \multicolumn{2}{c}{UESTC.} \\
        \cmidrule(lr){1-1} \cmidrule(lr){2-3} \cmidrule(lr){4-5} \cmidrule(lr){6-7}
        Num. of Classes  & \multicolumn{2}{c}{$5\times10$ stages} & \multicolumn{2}{c}{$5\times5$ stages} & \multicolumn{2}{c}{$4\times8$ stages}\\
        Classifier & CNN & NME & CNN & NME & CNN &  NME \\
        \cmidrule(lr){1-1} \cmidrule(lr){2-3} \cmidrule(lr){4-5} \cmidrule(lr){6-7}
        TCD\cite{park2021class} & 3.24 & 1.71 & 7.82 & 3.15 & 9.85 & 7.08 \\
        SNRO & \textbf{1.92} & \textbf{1.40} & \textbf{7.16} & \textbf{2.98} & \textbf{8.50} & \textbf{7.04} \\ 
        \hline
    \end{tabular}
    \vspace{0.5em}
    \caption{Average forgetting of SNRO and TCD on UCF101, HMDB51, and UESTC-MMEA-CL.}
    \label{tab:forgetten}
    \vspace{-2.0em}
\end{table}

\begin{table}
    \centering
    \begin{tabular}{ccccc}
        \hline
        \multicolumn{2}{c}{Examples Sparse} & \multirow{2}{*}{Early Break} & \multirow{2}{*}{CNN} & \multirow{2}{*}{NME} \\
        Sparse Extract & Frame Alignment &  & & \\
        \cmidrule(lr){1-2} \cmidrule(lr){3-3} \cmidrule(lr){4-5}
        \ding{53} & \ding{53} & \ding{53} & 72.06 & 73.05 \\
        \ding{53} & \ding{53} & \checkmark & 72.79 & 73.40 \\
        $T/2$ & Avg. & \checkmark & 74.69 & 74.17 \\
        $T/2$ & Dupe. & \checkmark & 75.33 & 75.13 \\
        $T/4$ & Avg. & \checkmark & 74.33 & 73.53 \\
        $T/4$ & Dupe. & \ding{53} & 75.24 & 75.53 \\
        $T/4$ & Dupe. & \checkmark & \textbf{76.12} & \textbf{75.97} \\
        \hline
    \end{tabular}
    \vspace{0.5em}
    \caption{Ablation study results about Examples Sparse $\&$ Early Break on UCF101. Avg. means average, Dube. means duplicate. The incremental stage is set to $5\times10$. The random seed is set to 1000.}
    \label{tab:ablation_1}
    \vspace{-1.0em}
\end{table}

\begin{table}
    \centering
    \begin{tabular}{cccccc}
        \hline
        Classifier & NoF. & NoV. & Memory Size & CNN & NME \\
        \cmidrule(lr){1-1} \cmidrule(lr){2-4} \cmidrule(lr){5-6}
        TCD & 8 & 5 & 6Mb & 72.06 & 73.05 \\
        TCD & 8 & 10 & 12Mb & 75.19 & \textbf{75.40}\\
        SNRO & 4 & 10 & 6Mb & \textbf{75.33} & 75.13 \\
        \hline
    \end{tabular}
    \vspace{0.5em}
    \caption{Ablation study results about SNRO and TCD Under the premise of the same memory set size. NoV. represents how many memory videos are stored for each class, and NoF. represents how many frames are stored in each video.}
    \label{tab:ablation_2}
    \vspace{-2.0em}
\end{table}

\subsubsection{Examples Sparse $\&$ Early Break}
In order to demonstrate the effectiveness of Examples Sparse and Early Break, we conducted experiments with different configurations on the UCF101 dataset.
Table \ref{tab:ablation_1} shows the experiments' results.
Examples Sparse brings a major performance promotion, mainly because it significantly increases the capacity of the memory set and suppresses high semantic features.
Early Break effectively prevents the tendency of over-fit to new classes, achieving a $0.73\%$ CNN improvement with the same memory set construction method.

\subsubsection{Less Spatio-Temporal Information Is Not Bad}
We fix the storage space and the number of memory samples respectively and conduct experiments on TCD and SNRO. 
Table \ref{tab:ablation_2} shows that when the fixed storage space is $6Mb$, the accuracy of SNRO is significantly better than that of TCD, mainly because SNRO can save twice samples for each old class and significantly reduce the forgetting of old classes. Even if the memory consumption of TCD is increased to $12Mb$, managing to maintain the same size memory sets as SNRO, SNRO still be competitive. Notice that under the same size memory sets, less spatio-temporal information does not bring performance loss to SNRO, and even surpasses TCD which consumes twice the storage space in terms of CNN performance.

\section{Conclusion}
We present a novel framework for video class-incremental learning named SNRO, which slightly shifts the features of new classes during their training stage and helps greatly improve the performance of old classes.
Specifically, we introduced Examples Sparse and Early Break, first one helps store more examples and forces the model to pay more attention to low-semantic features which are harder to be forgotten, and the second prevents the model from overstretching into the high semantic space of the current task.
Compared with existing works, SNRO achieves higher performance with the same memory consumption, even if its memory consumption is limited to half of the other works, SNRO is still competitive.


\bibliographystyle{acm}
\bibliography{Article}

\end{document}